\documentclass[letterpaper]{article} 
\usepackage{aaai2027}  
\usepackage[hyphens]{url}  
\usepackage{graphicx} 
\usepackage{natbib}  
\usepackage{caption} 
\usepackage{amsfonts} 
\usepackage{relsize} 
\usepackage{algorithm}
\usepackage{algorithmic}
\usepackage{algorithm}
\usepackage{algorithmic}

\usepackage{listings}
\usepackage{amsmath} 
\usepackage{multirow}
\usepackage{newfloat}
\usepackage{pifont}
\newcommand{\inc}[1]{\textsubscript{\textcolor{green}{\tiny #1}}}
\newcommand{\incr}[1]{\textsubscript{\textcolor{red}{\smaller{#1}}}}
\usepackage{relsize}
\usepackage{makecell}

\DeclareCaptionStyle{ruled}{labelfont=normalfont,labelsep=colon,strut=off} 
\floatstyle{ruled}
\newfloat{listing}{tb}{lst}{}
\floatname{listing}{Listing}

\usepackage{booktabs}
\usepackage{cleveref}
\usepackage{tabularx}
\usepackage[table]{xcolor}

\usepackage{newfloat}
\usepackage{listings}
\DeclareCaptionStyle{ruled}{labelfont=normalfont,labelsep=colon,strut=off} 
\floatstyle{ruled}
\newfloat{listing}{tb}{lst}{}
\floatname{listing}{Listing}

\usepackage{booktabs}

\title{BoT-Feedback: Grounding Multimodal Reasoning in Biomechanical Evidence for Explainable Human Action Feedback}
\author{
    Xu Dong\textsuperscript{\rm 1},
    Wanqing Li\textsuperscript{\rm 2},
    Anthony Adeyemi-Ejeye\textsuperscript{\rm 1},
    Andrew Gilbert\textsuperscript{\rm 1}
}
\affiliations{
    \textsuperscript{\rm 1}University of Surrey, Guildford, United Kingdom\\
    \textsuperscript{\rm 2}University of Wollongong, Wollongong, Australia\\
    x.dong@surrey.ac.uk, a.gilbert@surrey.ac.uk
}

\begin{document}

\maketitle
\begin{figure*}[htbp]
    \centering
    \includegraphics[width=1\textwidth]{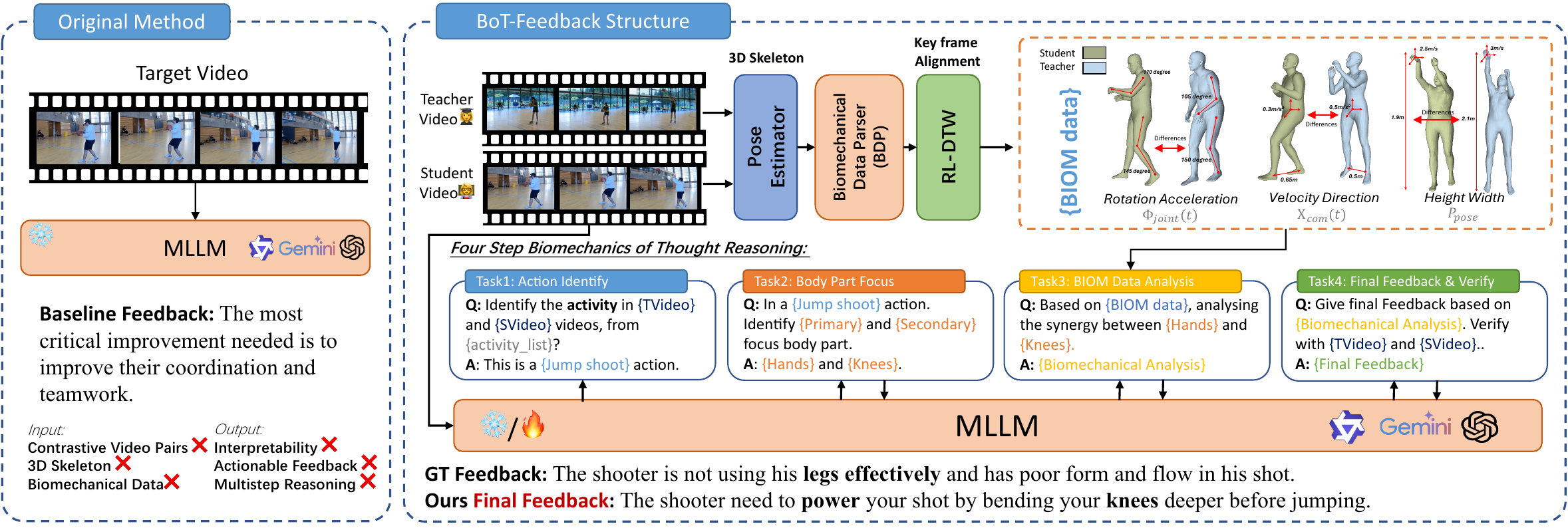}
    \caption{Overview of BoT-Feedback. A Biomechanical Data Parser extracts and aligns expert–student motion descriptors, which guide a four-stage Biomechanics of Thought process to generate evidence-grounded, actionable feedback.}
    \label{fig:teaser}
\end{figure*}

\begin{abstract}

Multimodal Large Language Models (MLLMs) have demonstrated impressive capabilities in visual understanding and multimodal reasoning, yet they remain fundamentally limited in Human Action Feedback Generation. Existing methods infer coaching feedback directly from visual observations, producing generic advice, limited interpretability, and physically implausible hallucinations. In contrast, expert human coaches diagnose performance through explicit biomechanical reasoning over joint kinematics, posture, and body dynamics. We introduce \textbf{BoT-Feedback}, a framework that grounds MLLM reasoning in structured biomechanical evidence. Our key contribution is Biomechanics of Thought (BoT), a four-stage reasoning framework that progressively identifies the action, localises the critical body regions, analyses quantitative biomechanical differences between expert and student performances, and synthesises interpretable coaching feedback. To support this reasoning process, we develop a plug-and-play Biomechanical Data Parser (BDP) that converts videos into structured biomechanical descriptors and an alignment strategy that temporally matches expert and student motions. We further introduce \textbf{BiomAF}, a benchmark containing paired teacher–student videos, 3D skeletons, biomechanical attributes, and expert-coaching annotations. Experiments across twelve open- and closed-source MLLMs demonstrate that grounding reasoning in biomechanical evidence consistently improves feedback quality, interpretability, and robustness while substantially reducing biomechanical hallucinations. BoT-Feedback improves the average expert evaluation score from 2.07 to 2.95 (+40\%), enabling compact open-source MLLMs to approach the performance of substantially larger proprietary systems for explainable action feedback generation.

\end{abstract}

Human Action Feedback Generation (HAFG) aims to produce actionable natural-language guidance to improve human movement, with applications in sports coaching, skill training and rehabilitation. Human experts do not diagnose movement quality from appearance alone. Instead, they compare student and expert executions using measurable biomechanical evidence, including joint angles, segment velocities, acceleration, posture, and body alignment \cite{barrios2010gait,hewett2005biomechanical,mcpherson1989relation,winter2009biomechanics,lees2002technique}. This diagnosis typically follows a structured process: identifying the action, determining the body segments most relevant to successful execution, comparing their motion quantitatively, and translating the resulting discrepancy into corrective advice. Such evidence-based reasoning makes professional feedback specific, interpretable, and explicitly grounded in structured biomechanical evidence.

Multimodal Large Language Models (MLLMs) \cite{li2023blip2bootstrappinglanguageimagepretraining,liu2023visualinstructiontuning,zhu2023minigpt4enhancingvisionlanguageunderstanding} have shown strong capabilities in visual understanding and cross-modal generation, but current systems generate action feedback primarily from visual appearance and linguistic priors \cite{li2024llavanextinterleavetacklingmultiimagevideo,chen2024internvlscalingvisionfoundation}. They lack an explicit intermediate physical representation over which structured reasoning can be performed. Consequently, their outputs are often disconnected from inspectable evidence, resulting in generic suggestions such as \textit{work on your form} rather than corrections tied to a measurable discrepancy, such as insufficient knee flexion before take-off. Moreover, without explicit joint trajectories, temporal alignment, or derived kinematic quantities, MLLMs may produce statements that are visually plausible but biomechanically inconsistent. This limits their interpretability and reliability for professional coaching and rehabilitation-oriented applications.

We argue that improving HAFG requires changing not only what information is given to an MLLM, but how the model is guided to reason over it. Rather than inferring coaching advice directly from visual observations, the model should reason through structured biomechanical evidence of movement. Inspired by expert diagnostic practice \cite{winter2009biomechanics,lees2002technique,simon1978information}, this representation should support a progression from action recognition to body-part selection to quantitative motion comparison to, finally, evidence-backed feedback.

To realise this idea, we introduce  \textbf{BoT-Feedback}, whose central component is \textbf{Biomechanics of Thought (BoT):} a four-stage reasoning framework that grounds each diagnostic step in biomechanical evidence. Conceptually, BoT replaces semantic-only intermediate reasoning with reasoning grounded in explicit, structured biomechanical evidence, allowing each intermediate conclusion to be traced back to measurable biomechanical evidence. As shown in \Cref{fig:teaser}, BoT first identifies the action performed, then selects the body regions most relevant to its kinetic chain, compares aligned student and expert biomechanical measurements, and finally generates concise feedback that is verified against visual observations. Unlike existing reasoning paradigms (Chain-of-Thought, Tree-of-Thought, Graph-of-Thought), which operate primarily in semantic token space, BoT constrains every reasoning step with measurable biomechanical evidence, making intermediate conclusions inspectable and reducing biomechanical hallucinations. 

To provide the structured biomechanical evidence required by BoT, we develop a plug-and-play \textbf{Biomechanical Data Parser (BDP)} that converts videos into interpretable descriptors of global body motion, local joint kinematics and posture. Because expert and student performance may differ in execution speed, we introduce \textbf{Robust Local Dynamic Temporal Warping (RL-DTW)} to establish temporally aligned biomechanical measurements prior to reasoning.

Existing benchmarks \cite{grauman2024egoexo4dunderstandingskilledhuman,ashutosh2025expertafexpertactionablefeedback} lack the combination of paired expert-student video, structured 3D motion representations, derived biomechanical attributes, and expert corrective feedback. We therefore introduce \textbf{BiomAF}, a biomechanically enriched benchmark constructed from EgoExo4D and ExpertAF. BiomAF contains 992 videos across six action categories, organised into curated teacher–student pairs and augmented with 3D skeletal trajectories, temporally aligned biomechanical descriptors, and expert feedback. The benchmark enables evaluation not only of final language quality, but also of whether generated advice is supported by observable physical discrepancies. 

We evaluate BoT-Feedback across twelve open- and closed-source MLLMs and six action categories. Grounding MLLM reasoning in biomechanical evidence improves the average evaluation score from 2.07 to 2.95, with particularly large gains for compact open-source models. On Qwen3-VL-8B, for example, BoT-Feedback raises the overall score from 2.15 to 3.59. Stepwise ablations show that the largest improvement occurs when the model is given aligned quantitative biomechanical evidence, while robustness experiments demonstrate that the framework remains effective under jitter or noise in pose estimation and action identification. These results support the central hypothesis that structured physical reasoning produces more precise and reliable feedback than direct inference from visual observations.

\noindent Our main contributions are:

\begin{itemize}
    \item We introduce \textbf{Biomechanics of Thought (BoT)}, a four-stage reasoning framework that grounds MLLM inference in explicit biomechanical evidence, enabling interpretable and actionable human action feedback.
    \item We develop a plug-and-play biomechanical evidence pipeline comprising a \textbf{Biomechanical Data Parser} and an \textbf{RL-DTW} alignment, which transforms expert and student videos into structured, temporally corresponding motion descriptors.
    \item We introduce \textbf{BiomAF}, a benchmark of curated teacher–student action pairs augmented with 3D skeletons, biomechanical attributes, and expert feedback, and demonstrate consistent improvements across twelve MLLMs in zero-shot and fine-tuned settings
    
\end{itemize}

\section{Related Work}
\paragraph{Human Skill Understanding}

Human Skill Understanding has evolved from action quality assessment (AQA) to MLLM-based feedback reasoning. Early methods \cite{pirsiavash2014assessing, parmar2019action, parmar2017learning, xu2019learning, xiang2018s3d, Doughty_2019_CVPR,xu2022likert, xu2024fineparser, Zhou_2024} regress scalar scores from spatiotemporal features,  while recent work improve interpretability \cite{dong2026uila, hou2025interpretable, han2025finecausal} and generate natural language coaching with MLLMs \cite{xu2025languageguided, qi2025explainable, xu2025qualityguided}. However, these methods remain grounded in generic visual features rather than explicit biomechanical reasoning, motivating our Biomechanics of Thought framework.

\paragraph{Multimodel Reasoning}
MLLMs~\cite{qwen3vl2025, comanici2025gemini25pushingfrontier, lu2024deepseek, damon2024videollama3, lin2023video} have achieved human-level reasoning across diverse benchmarks. Existing paradigms such as CoT~\cite{wei2023chainofthoughtpromptingelicitsreasoning}, ToT~\cite{treeofthought}, GoT~\cite{Besta_2024}, PoT~\cite{chen2023programthoughtspromptingdisentangling}, and VoT~\cite{fei2024videoofthoughtstepbystepvideoreasoning} enhance logical depth but operate purely in semantic token space, making them prone to \emph{biomechanical hallucinations} — inferences that are linguistically plausible but kinematically infeasible. In contrast, our Biomechanics of Thought grounds each reasoning step in quantitative biomechanical features and physical verification, anchoring MLLM reasoning in real-world motion dynamics.

\paragraph{Physics-informed Action Analysis}

Vision-based methods are often limited by image noise and occlusion, failing to capture the underlying physical dynamics of human actions. Recent works integrate physical constraints into vision pipelines: PhysCap~\cite{shimada2020physcapphysicallyplausiblemonocular} enforces gravity and contact constraints for monocular 3D capture; SIP~\cite{vonmarcard2017sparseinertialposerautomatic} combines IMU sensors with physics optimisation for skeletal reconstruction; CloseInt~\cite{Huang_2024_CVPR} applies proxemics and physics-guided adaptation for multi-person interactions. However, none of these methods translate physical metrics into language-level coaching instructions. We bridge this gap by leveraging biomechanical data to enhance MLLM interpretability in understanding kinematic causalities underlying human movement.

\section{Methodology}
BoT-Feedback is centred on Biomechanics of Thought (BoT), a structured reasoning framework (shown in \Cref{fig:teaser}) that grounds MLLM inference in explicit biomechanical evidence, rather than reasoning directly from visual observations. To support this reasoning process, we first construct structured biomechanical representations from expert and student videos using a \textbf{Biomechanical Data Parser (BDP)} and a temporal alignment module. \textbf{BoT} then performs multi-stage reasoning over these representations to generate interpretable and actionable coaching feedback.

Rather than inferring feedback directly from visual observations, expert coaches \cite{winter2009biomechanics,lees2002technique,simon1978information} progressively narrow their reasoning from global action recognition to local biomechanical diagnosis before producing advice. By explicitly encoding this expert workflow into a four-step pipeline, Biomechanics of Thought ensures that MLLM reasoning remains grounded, interpretable, and aligned with professional biomechanical standards.


\subsection{Biomechanical Data Parser}
The role of the BDP is not to replace the visual encoder but to expose an explicit physical representation over which BoT can reason. Rather than forcing the MLLM to infer biomechanics implicitly from pixels, the BDP provides structured biomechanical evidence describing global motion, local joint kinematics, and posture. It first applies SAM3D~\cite{sam3dteam2025sam3d3dfyimages} to lift 2D frames into a 3D skeletal coordinate system, yielding joint positions $P \in \mathbb{R}^{F \times J \times 3}$, with a person-selection method to identify the primary subject.

\subsubsection{Torso Normalisation} To ensure the model remains invariant to the subject's physical size or camera distance, the BDP calculates a normalisation factor based on torso length (the distance between the neck and the pelvis). All subsequent spatial metrics—such as velocities and heights—are scaled relative to this reference, allowing the framework to generalise across different body shapes. 

\subsubsection{Biomechanical Feature Construction} 
\begin{figure}[t]
    \centering
    \caption{The Biomechanical Data Parser (BDP) converts raw 3D skeletons into structured biomechanical descriptors for Biomechanics of Thought.}
    \includegraphics[width=1\linewidth, trim={0 0.2cm 0 0}, clip]{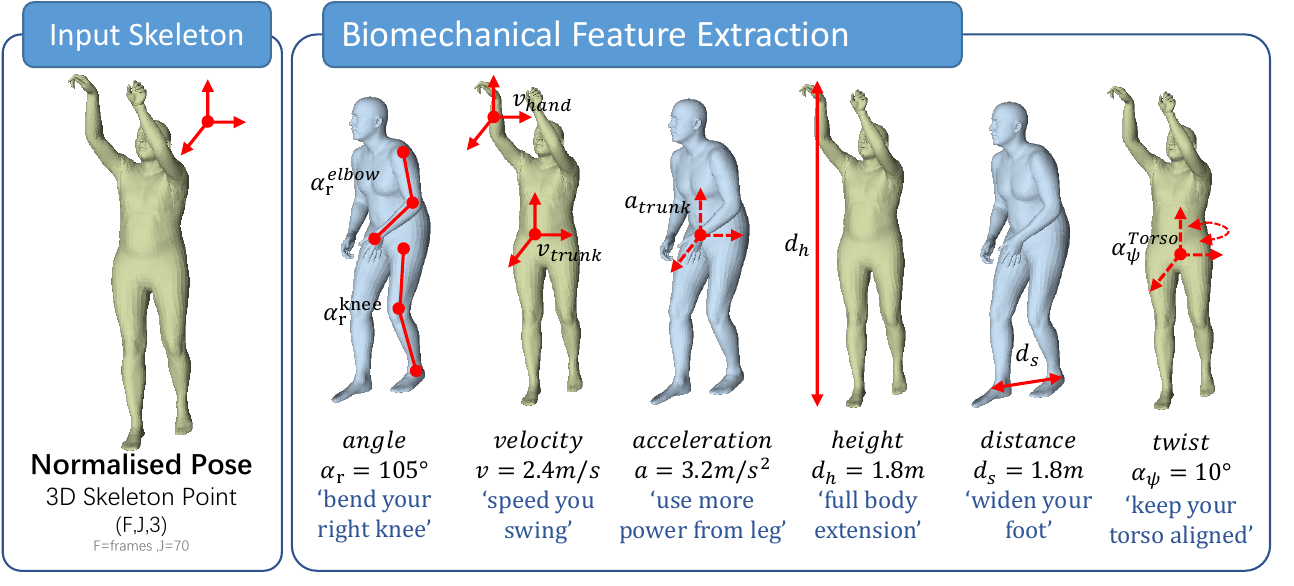}
    \label{fig:human}
\vspace{-15pt}
\end{figure}

Inspired by classical biomechanics \cite{winter2009biomechanics, mcpherson1989relation}, the BDP transforms raw skeletons into intuitive structured biomechanical descriptors (\Cref{fig:human}) that form the reasoning space for BoT. The representation comprises three complementary levels:

\noindent\textbf{\textit{Global Body Motion}}, $\mathbf{X}_{CoM}(t)$ models whole-body dynamics using the torso Centre of Mass (CoM), from which velocity and acceleration are estimated and temporally smoothed using a Savitzky--Golay filter \cite{1964AnaCh}.


\noindent\textbf{\textit{Local Joint Kinematics}}: \textbf{\textit{Local Joint Kinematics}} $\mathbf{\Phi}_{joint}(t)$ captures joint-level motion through flexion angles, velocities, accelerations, and movement directions for key anatomical segments.


\noindent \textbf{\textit{Posture Configuration Features}} $\mathbf{P}_{pose}$ describe global posture, including base of support, reachability, and torso orientation.
Together, these descriptors form a multi-scale biomechanical profile,
$$\mathbf{P}_{bio}(t) = \left[ \mathbf{X}_{CoM}(t), \mathbf{\Phi}_{joint}(t), \mathbf{P}_{pose} \right]^T$$ 
which provides the structured biomechanical evidence for Biomechanics of Thought. Detailed formulations are provided in the Supplementary Material.

\subsubsection{Robust Local Dynamic Time Warping (RL-DTW)}
Because expert and student executions naturally differ in speed, corresponding biomechanical events rarely occur at identical timestamps. RL-DTW establishes temporally corresponding biomechanical measurements prior to reasoning, ensuring that BoT compares equivalent phases of the movement.

Dynamic Time Warping (DTW) \cite{Sakoe1978DynamicPA} is a non-linear alignment operator that aligns temporal variations between two multidimensional trajectories. Given sequences $\mathbf{S} = \{s_1, \dots, s_{\tau_1}\}$ and $\mathbf{Q} = \{q_1, \dots, q_{\tau_2}\}$, the vanilla DTW minimizes the cumulative distance:$(\mathcal{P}^*, \mathcal{D}_{DTW}) = \text{DTW}(\mathbf{S}, \mathbf{Q})$, where $\mathcal{P}^*$ is the optimal warping path and $\mathcal{D}_{DTW}$ is the global minimum cost. However, vanilla DTW fails on skeleton data due to low feature discriminability and high noise sensitivity, which cause blurred alignments and path drift. Since BoT requires temporally corresponding biomechanical measurements, we introduce  Robust Local DTW (RL-DTW). Instead of skeleton coordinates, we employ normalised biomechanical features from our Biomechanical Data Parser to unify heterogeneous scales. These features are reconstructed into local temporal blocks via a sliding block to encapsulate short-term action context:$\hat {v}_t = [v_t, v_{t+n}]$, where $n$ is the temporal offset. Finally, the optimal alignment is determined by computing the warping path through the block sequences $\hat{\mathcal{V}}_1$ and $\hat{\mathcal{V}}_2$, where $\hat{\mathcal{V}} = \{\hat{v}_t\}$:$$(\mathcal{P}^*, \mathcal{D}_{RL}) = \text{DTW}(\hat{\mathcal{V}}_1, \hat{\mathcal{V}}_2)$$ By aligning these higher-dimensional temporal blocks, RL-DTW yields a more stable cost matrix, ensuring that the resulting distance metric is robust to pose estimation noise and inherent motion variability and provides a high-confidence measure of action similarity.

\begin{figure}[th]
  \centering
  \caption{The proposed four-step Biomechanics of Thought framework for progressive reasoning. Detailed prompt templates and multi-step dialogue examples can be found in the Supplementary Material.}
  \vspace{3pt}
  \includegraphics[width=1\linewidth, trim={0 0.1cm 0 0}, clip]{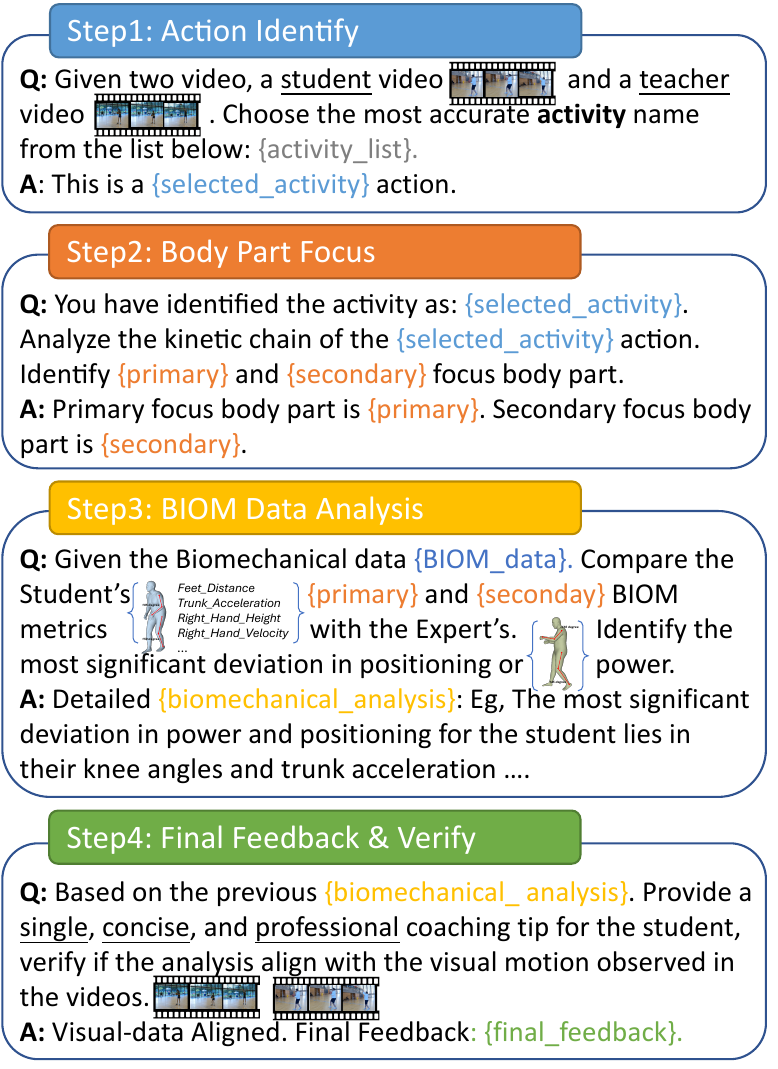}
  \label{fig:bot}
  \vspace{-20pt}
\end{figure}

\subsection{Biomechanics of Thought Reasoning}
Biomechanics of Thought (BoT) is the central reasoning framework of BoT-Feedback. Rather than producing feedback directly from visual features,the gradual improvement across the four stages suggests that each stage contributes complementary information: action identification establishes task context, body-part focus localises the relevant anatomy, biomechanical analysis provides quantitative evidence, and verification ensures consistency between the evidence and the generated feedback. (\Cref{fig:bot}). 

Existing reasoning frameworks (CoT, ToT, GoT, VoT) \cite{barez2025chain, stechly2025chainthoughtlessnessanalysiscot} improve reasoning by manipulating semantic reasoning trajectories. Unlike these methods, which improve reasoning through semantic decomposition, BoT introduces domain-grounded reasoning over structured biomechanical evidence. Consequently, each intermediate conclusion is directly supported by measurable biomechanical evidence rather than latent semantic associations.

\subsubsection{\textbf{Step 1: Action Identification}}
This stage takes two videos as input to perform a classification task. By identifying the specific action category $\mathtt{\{selected\_activity}\}$  from a predefined activity list $\mathtt{\{activity\_list}\}$ (e.g., a Mikan Layup), the model establishes the task-specific biomechanical context. This foundational step ensures that all subsequent stages of reasoning align with the professional-technical standards for the identified action. Note that $\mathtt{\{activity\_list\}}$ is a predefined candidate set; the model selects purely from visual inference without any ground-truth supervision.
\subsubsection{\textbf{Step 2: Body Part Focus}}
In this stage, based on the previous $\mathtt{\{selected\_activity}\}$, the reasoning process transitions from global observation to task-specific body parts. By inferring the corresponding $\mathtt{\{primary\}}$ and $\mathtt{\{secondary\}}$ body regions (e.g., the elbow or knee), this step facilitates a deeper analysis of the kinetic chain. This refinement step effectively eliminates irrelevant visual noise and establishes a precise focus for the subsequent biomechanical data analysis. Likewise, $\mathtt{\{primary\}}$ and $\mathtt{\{secondary\}}$ are selected from a fixed biomechanical candidate set without oracle labels.
\subsubsection{\textbf{Step 3: Biomechanical Data Analysis}}
This is the core reasoning stage of BoT; the model performs explicit quantitative reasoning using biomechanical data $\mathtt{\{BIOM\_data\}}$ aligned across expert and student data. By extracting biomechanical features associated with the $\mathtt{\{primary\}}$ and $\mathtt{\{secondary\}}$ body parts identified in previous steps, the model conducts an objective $\mathtt{\{biomechanical\_analysis\}}$. To exploit pretrained reasoning capabilities of the MLLM, this step detects precise deviations between the student and the expert and diagnoses the underlying biomechanical causes.

\begin{table*}[thbp]
\centering
\caption{Comparison with 12 MLLMs on six action tasks. Higher is better except STD. \textbf{bold} denote the best performance, and \underline{underlined} denote the second-best. Models marked with $^\dagger$ indicate that an image sequence is used as input. Action tasks are defined as: SP (Soccer Penalty), SD (Soccer Dribbling), BRL (Basketball Reverse Layup), BM (Basketball Mikan Layup), BJS (Basketball Jump Shot), and BO (Bouldering).}
\footnotesize
\label{tab:full_results_with_bouldering}
\setlength{\tabcolsep}{2pt} 
\begin{tabular}{l | cc | cc | cc | cc | cc | cc || cc | c}
\toprule
\multirow{2}{*}{\textbf{Model}} & \multicolumn{2}{c|}{\textbf{SP$\uparrow$}} & \multicolumn{2}{c|}{\textbf{SD$\uparrow$}} & \multicolumn{2}{c|}{\textbf{BRL$\uparrow$}} & \multicolumn{2}{c|}{\textbf{BM$\uparrow$}} & \multicolumn{2}{c|}{\textbf{BJS$\uparrow$}} & \multicolumn{2}{c||}{\textbf{BO$\uparrow$}} & \multicolumn{2}{c|}{\textbf{Overall Avg.$\uparrow$}} & \textbf{STD$\downarrow$} \\
\cmidrule(lr){2-3} \cmidrule(lr){4-5} \cmidrule(lr){6-7} \cmidrule(lr){8-9} \cmidrule(lr){10-11} \cmidrule(lr){12-13} \cmidrule(lr){14-15} \cmidrule(lr){15-16}
& Base & \textbf{Ours} & Base & \textbf{Ours} & Base & \textbf{Ours} & Base & \textbf{Ours} & Base & \textbf{Ours} & Base & \textbf{Ours} & Base & \textbf{Ours} & \textbf{Ours}\\
\midrule\midrule
\rowcolor{gray!15} \multicolumn{16}{l}{\textbf{Open-Source Models}} \\ 
Video-ChatGPT & 1.98 & 1.90 & 1.45 & 1.70 & 1.79 & 2.08 & 1.84 & 2.14 & 1.65 & 2.03 & 1.63 & 1.62 & 1.72 & 1.91 \inc{+0.19} & 0.25 \\
VideoLLaVA-7B       & 1.65 & 2.76 & 1.86 & 1.80 & 1.59 & 1.84 & 1.51 & 2.27 & 1.49 & 1.77 & 2.06 & 2.03 & 1.69 & 2.08 \inc{+0.39} & 0.51 \\
VideoLLaMA3-7B   & 2.22 & 2.86 & 2.23 & 2.99 & 1.51 & 2.74 & 1.62 & 2.70 & 1.90 & 2.57 & 1.99 & 2.45 & 1.91 & 2.72 \inc{+0.81} & 0.29 \\
InternVideo2.5-8B  & 1.75 & 2.15 & 1.77 & 2.79 & 1.74 & 3.30 & \textbf{2.13} & 3.08 & 1.91 & 3.06 & 1.91 & 2.45 & 1.87 & 2.81 \inc{+0.94} & 0.48 \\
VideoChat-R1-7B   & 1.97 & 3.13 & 2.31 & 3.05 & 1.81 & 2.79 & 1.97 & 2.91 & 1.96 & 3.10 & 1.89 & 2.33 & 1.99 & 2.89 \inc{+0.90} & 0.42 \\
Qwen2.5-VL-7B        & 1.61 & 3.56 & \textbf{2.38} & 2.99 & 1.77 & 3.15 & 1.86 & 2.99 & 1.98 & 3.22 & 2.07 & 2.50 & 1.95 & 3.07 \inc{+1.12} & 0.31 \\
InternVL2.5-8B$^\dagger$   & 2.14 & 2.78 & 1.66 & \textbf{3.38} & 1.60 & 3.55 & 1.83 & 2.99 & 2.03 & 3.40 & 1.94 & 3.06 & 1.87 & 3.19 \inc{+1.32} & 0.34 \\
Qwen3-VL-8B        & \textbf{2.30} & 2.98 & 2.23 & 3.31 & \textbf{2.07} & \textbf{4.04} & 2.01 & \textbf{4.07} & \textbf{2.10} & \textbf{3.83} & 2.21 & \textbf{3.32} & \textbf{2.15} & \textbf{3.59} \inc{+1.44} & 0.50 \\
\midrule
\rowcolor{gray!15} \multicolumn{16}{l}{\textbf{Closed-Source Models}} \\
GPT-4o-mini$^\dagger$          & 1.97 & 3.37  & 2.19 & 2.80 & 1.61 & 3.17 & 1.93 & 3.38 & 1.95 & 3.24 & 2.17 & 2.32 & 1.97 & 3.05 \inc{+1.08} & 0.55 \\
Gemini 2.5 Flash   & 2.98 & \textbf{3.61} & 2.95 & \textbf{3.41} & 2.88 & 3.43 & \textbf{3.32} & 3.54 & \textbf{3.00} & 3.46 & \textbf{3.33} & 3.25 & 3.08 & 3.46 \inc{+0.38} & 0.28 \\
Gemini 3.1 Pro    & \textbf{3.45} & 3.54  & \textbf{3.46} & 3.21 & \textbf{3.48} & \textbf{4.23} & 2.96 & \textbf{4.11} & 2.92 & \textbf{4.08} & 3.21 & \textbf{4.08} & \textbf{3.25} & \textbf{3.88} \inc{+0.63} & 0.43 \\
\bottomrule
\end{tabular}
\end{table*}

\subsubsection{\textbf{Step 4: Verification and Final Feedback Generation}}
In this final stage, the $\mathtt{\{biomechanical\_analysis\}}$ are synthesised into concise, actionable coaching feedback $\mathtt{\{final\_feedback\}}$. The generated coaching advice is verified against both the original visual observations and the underlying biomechanical evidence before being returned to the user.

\noindent\textit{Training Objective:} Importantly, BoT does not require supervision of its intermediate reasoning steps. Only the final coaching feedback is supervised (Step 4), while the intermediate reasoning emerges from the structured biomechanical representation and the pretrained reasoning capabilities of the MLLM.


\section{Experiments}
In this section, we evaluate BoT-Feedback on \textit{BiomAF}, our biomechanical action feedback benchmark, covering implementation details, quantitative comparisons against state-of-the-art MLLMs, and qualitative analysis.
\subsubsection{BiomAF Benchmark }
The \textit{BiomAF} Benchmark builds upon two existing datasets: EgoExo4D~\cite{grauman2024egoexo4dunderstandingskilledhuman}, a large-scale multi-view dataset with 1,087 hours of video across nine activity categories with expert feedback annotations; and ExpertAF~\cite{ashutosh2025expertafexpertactionablefeedback}, which restructures EgoExo4D into expert-student contrastive pairs across eight sports sub-domains.
\subsubsection{The BiomAF Benchmark Construction}
BiomAF comprises 992 teacher–student video pairs spanning six action categories, augmented with 3D skeletons, biomechanical descriptors and expert coaching feedback. It is derived from EgoExo4D and ExpertAF after removing inconsistent pairs and extracting biomechanical features using SAM3D Body~\cite{yang2026sam3dbodyrobust}. Additional construction details are provided in the Supplementary Material.


\subsection{Implementation Details}  
All experiments were conducted on a single NVIDIA A100 GPU. Unless otherwise stated, videos were uniformly sampled and processed using SAM3D Body. Further implementation details are provided in the Supplementary Material.


\subsection{Baselines and Evaluation Metrics}
We evaluate our framework against a range of SoTA MLLMs, including both open-source and closed-source models. For open-source models, we include VideoLLaVA \cite{lin2023video}, Video-ChatGPT \cite{maaz2023video}, InternVideo2.5 \cite{wang2025internvideo25empoweringvideomllms}, VideoLLaMA3 \cite{damon2024videollama3}, VideoChat-R1 \cite{li2025videochatr1}, and the Qwen2.5-VL / Qwen3-VL series \cite{qwen3vl2025}, all of which support native video input. For models designed for image input, such as DeepSeek-VL2 \cite{lu2024deepseek} and InternVL2.5 \cite{chen2024internvl2}, we feed them sampled image sequences for comparison. For closed-source models, we evaluate Gemini 2.5 Flash \cite{comanici2025gemini25pushingfrontier}, Gemini 3.1 Pro \cite{team2026gemini3}, and GPT-4o-mini \cite{openai2024gpt4o} (Image sequence as input).
Traditional metrics such as BLEU and ROUGE are inadequate for evaluating action feedback, as they fail to capture biomechanical correctness. Therefore, following the recent trend of \emph{LLM-as-a-judge} evaluation \cite{liu2023gevalnlgevaluationusing, zhou2025physvlmenablingvisuallanguage, maaz2023video, fei2024videoofthoughtstepbystepvideoreasoning}, we employ a teacher-level MLLM to provide a more expert-aligned evaluation,  Gemini 3.0 \cite{team2026gemini3}. We define \textbf{four specialised metrics} for generated action feedback evaluation: Corrective Utility (CU): Actionable physical corrections. (1: Vague; 5: Precise); Expert Alignment (EA): Consistency with Ground Truth (GT) intent. (No penalty for more specific/correct feedback); Hallucination Check (HC): Physical hallucination check. (5: No error; 1: Physically impossible); and Information Density (ID): Presence of technical keywords vs. filler. (1: High fluff; 5: Dense insights). By comparing the Ground Truth (GT) with our generated feedback, the evaluator assigns a score ranging from 1 (worst) to 5 (best) for each metric.

\subsection{Comparison with SOTA}
To evaluate the generalisation and performance of the proposed framework, we conducted a comprehensive benchmark in a zero-shot setting. We selected 12 representative Multimodal Large Language Models (MLLMs) as baselines spans both Open- and Closed-Source models. As illustrated in \Cref{tab:full_results_with_bouldering}, we compared the baseline performance of these MLLMs against that when integrated with our BoT-Feedback pipeline. Grounding reasoning in structured biomechanical evidence consistently improves performance across all twelve evaluated MLLMs, regardless of architecture or training paradigm. Particularly large gains are observed for compact open-source models, suggesting that explicit physical reasoning can compensate for limited model capacity. For instance, it elevates Qwen3-VL-8B \cite{qwen3vl2025} overall average score from 2.15 to 3.59 ($\sim$67\% improvement). It makes the 8B open-source model comparable to large closed-source models \cite{comanici2025gemini25pushingfrontier,openai2024gpt4o} in its vanilla state. This effectively lowers the barrier for deploying professional action feedback generation systems in resource-constrained environments. The consistently low STD further confirms the robustness of our mechanism. We further investigate the effectiveness of fine-tuned BoT-Feedback in the Supplementary Material.

\subsubsection{Pilot Study}
To validate that our automated evaluation metrics align with human intuition, we conduct a pilot user study with 15 participants spanning novice, intermediate, and expert levels across diverse sports backgrounds. Participants are presented with anonymised feedback from three systems — Baseline MLLM, BoT-Feedback without BDP, and BoT-Feedback (Ours) — alongside paired student–expert videos, and evaluate each feedback across four dimensions (CU, EA, HC, ID) by selecting the best option before indicating their overall preference. As shown in \Cref{tab:user_study}, BoT-Feedback is consistently and significantly preferred over both baselines (57.3\%) across all four dimensions, a result that closely mirrors our automated MLLM evaluation, confirming the reliability of our proposed metrics as proxies for human expert judgment.

\begin{table}[htbp]
\footnotesize
\centering
\caption{User Study Results. \textbf{Percentage of participants selecting each system as best} across four evaluation dimensions and overall preference.}
\label{tab:user_study}
\resizebox{0.9\columnwidth}{!}{%
\begin{tabular}{@{}l c c c c c@{}}
\toprule
\textbf{System} & \textbf{CU} & \textbf{EA} & \textbf{HC} & \textbf{ID} & \textbf{Overall} \\
\midrule
Baseline MLLM       & 10.7\% & 12.0\% & 29.3\% & 1.3\% & 12.0\% \\
BoT (w/o BDP)       & 32.0\%   &28.0\%   & 21.3\%  & 26.7\%    & 30.7\% \\
BoT-Feedback (Ours) & \textbf{57.3\%}  & \textbf{60.0\%} &  \textbf{49.3\%} &  \textbf{72.0\%}   &  \textbf{57.3\%} \\
\bottomrule
\end{tabular}%
}
\end{table}

\subsection{Analysis of the BoT-Feedback Reasoning Process}

To understand how each reasoning stage contributes to feedback generation, we incrementally enable the four stages of Biomechanics of Thought. This analysis tests our central hypothesis that progressively grounding reasoning in structured biomechanical evidence leads to more accurate and interpretable coaching feedback. \Cref{tab:botablation} illustrates the performance improvement across 6 actions. 
The progressive improvement across all four stages validates the central hypothesis underlying BoT: decomposing coaching into interpretable reasoning stages enables increasingly accurate diagnostic feedback. \textbf{Step 3 (Biomechanical Data Analysis)} contributes the largest single improvement (\textbf{+0.83}), demonstrating that the principal gains arise from reasoning over explicit biomechanical evidence rather than from simply decomposing the task into multiple prompting stages. Finally, \textbf{Step 4 (Verification and Final Feedback Generation)} achieves a peak of 3.07, demonstrating that the verification mechanism can rectify previous misjudgments and refine the final output into accurate, professional feedback. 

To illustrate, consider the feedback generation for a basketball jump shot: while the Baseline provides only a generic instruction: \textit{“Ensure a complete and high follow-through”}, integrating Step 2 (Body Part Focus) allows the model to identify the key power source: \textit{“Remember to use your legs for power.”} However, it is Step 3 (Biomechanical Data Analysis) that yields the most precise, actionable correction: \textit{“Dip deeper by bending your knees more before your jump shot to generate more power.”}. This progression illustrates how BoT moves from generic semantic guidance to anatomically targeted advice and finally to quantitative, actionable coaching grounded in measurable movement characteristics.
\begin{table}[h]
\centering
\caption{Performance of incremental BoT reasoning stages. Subscripts denote gains over the baseline.}
\label{tab:botablation}
\resizebox{\columnwidth}{!}{
\begin{tabular}{l|ccccc}
\toprule
\textbf{Actions} & \textbf{Baseline $\uparrow$} & \textbf{Step 1 $\uparrow$} & \textbf{Step 2 $\uparrow$} & \textbf{Step 3 $\uparrow$} & \textbf{Step 4 $\uparrow$} \\ \midrule\midrule
SP  & 1.61 & 2.01\inc{+0.40} & 2.09\inc{+0.48} & 2.91\inc{+1.30} & 3.56\inc{+1.95} \\
SD  & 2.38 & 2.84\inc{+0.46} & 2.60\inc{+0.22} & 2.95\inc{+0.57} & 2.99\inc{+0.61} \\
BRL & 1.77 & 2.10\inc{+0.33} & 2.38\inc{+0.61} & 2.76\inc{+0.99} & 3.15\inc{+1.38} \\
BM  & 1.86 & 2.29\inc{+0.43} & 2.62\inc{+0.76} & 2.99\inc{+1.13} & 2.99\inc{+1.13} \\
BJS & 1.98 & 2.34\inc{+0.36} & 2.58\inc{+0.60} & 2.62\inc{+0.64} & 3.22\inc{+1.24} \\
BO  & 2.07 & 2.10\inc{+0.03} & 2.05\incr{-0.02} & 2.46\inc{+0.39} & 2.50\inc{+0.43} \\ \midrule
\textbf{Avg.} & \textbf{1.95} & \textbf{2.28\inc{+0.33}} & \textbf{2.39\inc{+0.44}} & \textbf{2.78\inc{+0.83}} & \textbf{3.07\inc{+1.12}} \\ \bottomrule
\end{tabular}
}
\end{table}

\subsubsection{Comparison of different DTW methods}
To ensure a fair and consistent comparison of Biomechanical features, we evaluate RL-DTW's performance on a manually annotated alignment subset. For each teacher-student video pair, we define and label specific action keyframes, such as the "dribbling point" in football or the "release moment" in basketball, as ground truth for temporal correspondence. We conduct a comparative analysis based on three primary alignment metrics: Mean Absolute Error (MAE), Max Offset (maximum temporal deviation across the sequence), and Keyframe Error (alignment displacement at action keyframes). As shown in \Cref{tab:dtw_ablation}, our RL-DTW method demonstrates superior overall alignment performance, significantly reducing the Total Cost (2.28) compared to existing approaches. \Cref{fig:dtw_comparison} further illustrates the DTW alignment paths. A comparative analysis shows that our proposed RL-DTW path (orange solid) is perfectly aligned with the GT action keyframes (red), whereas other methods exhibit temporal drift. These results demonstrate that accurate temporal correspondence is essential for reliable biomechanical reasoning, since comparisons between misaligned movement phases directly degrade the quality of downstream physical analysis.

\begin{table*}[htbp] 
    \centering
    \
    \begin{minipage}[c]{0.66\textwidth} 
        \centering
        \caption{Ablation across different DTW configurations, including Vanilla DTW~\cite{Sakoe1978DynamicPA}, Derivative DTW~\cite{keogh2001derivative}, and Soft-DTW~\cite{cuturi2018softdtwdifferentiablelossfunction}, on alignment accuracy. The symbols $\downarrow$ indicate that lower values represent better performance. The Total Cost score is calculated by the weighted sum based on each metric.}
        \label{tab:dtw_ablation}
        \small 
        \begin{tabular}{l c cccc}
    \toprule
    \textbf{Configuration} & \textbf{Modality} & \textbf{MAE $\downarrow$} & \textbf{Max Offset $\downarrow$} & \textbf{Kframe Error $\downarrow$} & \textbf{Total Cost $\downarrow$} \\
    \midrule\midrule
    Vanilla DTW  & 3D Skeleton & 1.9064 & 4.0000 & 2.1210 & 3.01 \\
    Vanilla DTW  & BIOM & 1.6596 & 3.8387 & 1.8548 & 3.15 \\
    Derivative DTW  & 3D Skeleton & 1.7121 & 3.6774 & \textbf{1.4194} & 2.63 \\
    Soft-DTW & BIOM Features & 1.6351 & 3.8387 & 1.8387 & 2.79 \\ 
    \midrule
    \textbf{RL-DTW (Ours)} & \textbf{BIOM} & \textbf{1.4148} & \textbf{3.1290} & 1.4960 & \textbf{2.29} \\
    \bottomrule
\end{tabular}
    \end{minipage}
    \hfill
    \begin{minipage}[c]{0.27\textwidth} 
        \centering
        \captionof{figure}{DTW alignment paths visualisation. Red points denote the Ground Truth matching points.} 
        \includegraphics[width=0.9\linewidth]{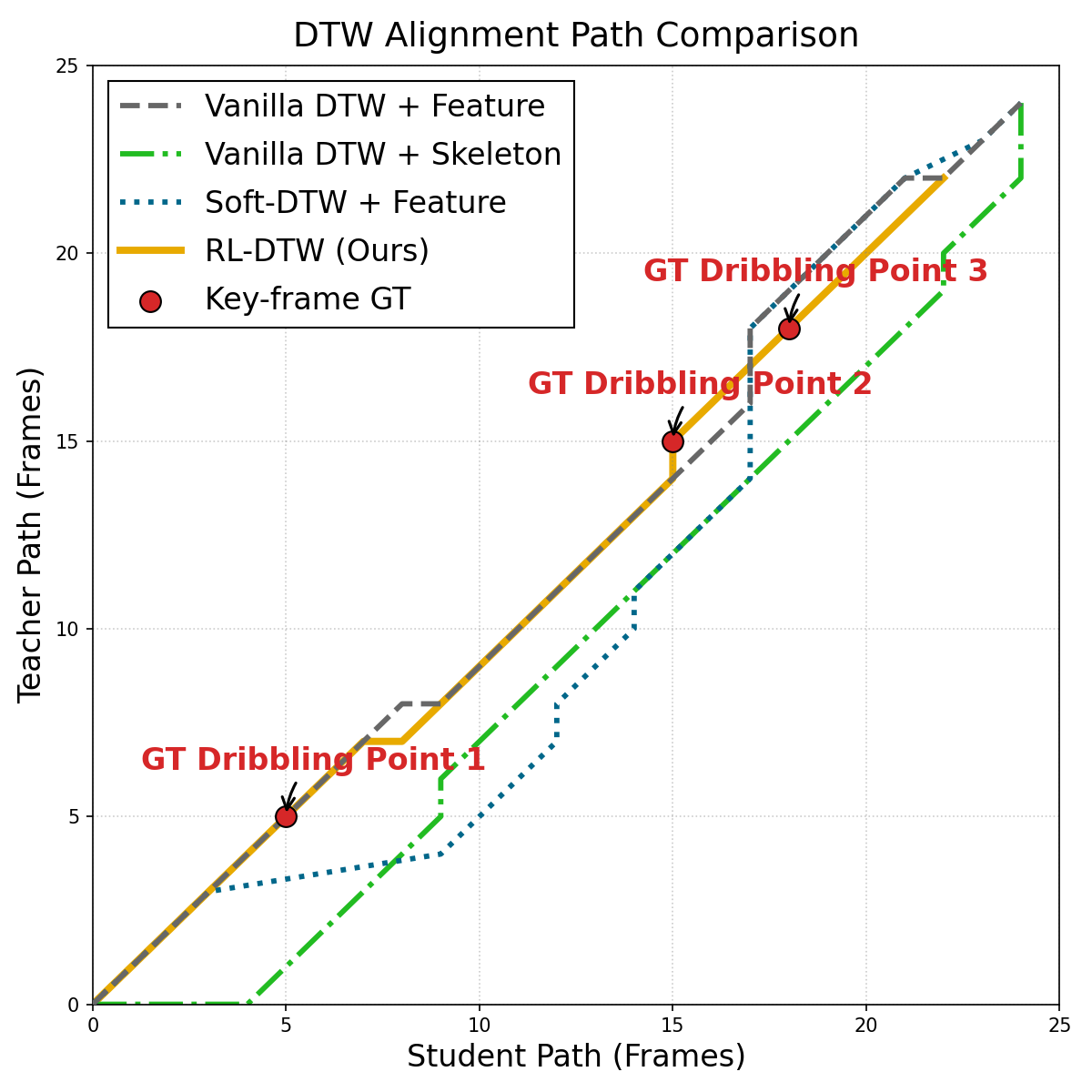}
        \label{fig:dtw_comparison}
    \end{minipage}
    \vspace{-10pt}
\end{table*}



\begin{figure}[!t]
    \centering
    \caption{Qualitative comparison between baseline MLLMs and BoT-Feedback.  \underline{\textbf{Bold underlined text}} highlights key biomechanical diagnoses.}
    \includegraphics[width=1.0\linewidth, trim={0.3cm 3cm 0.2cm 0}, clip]{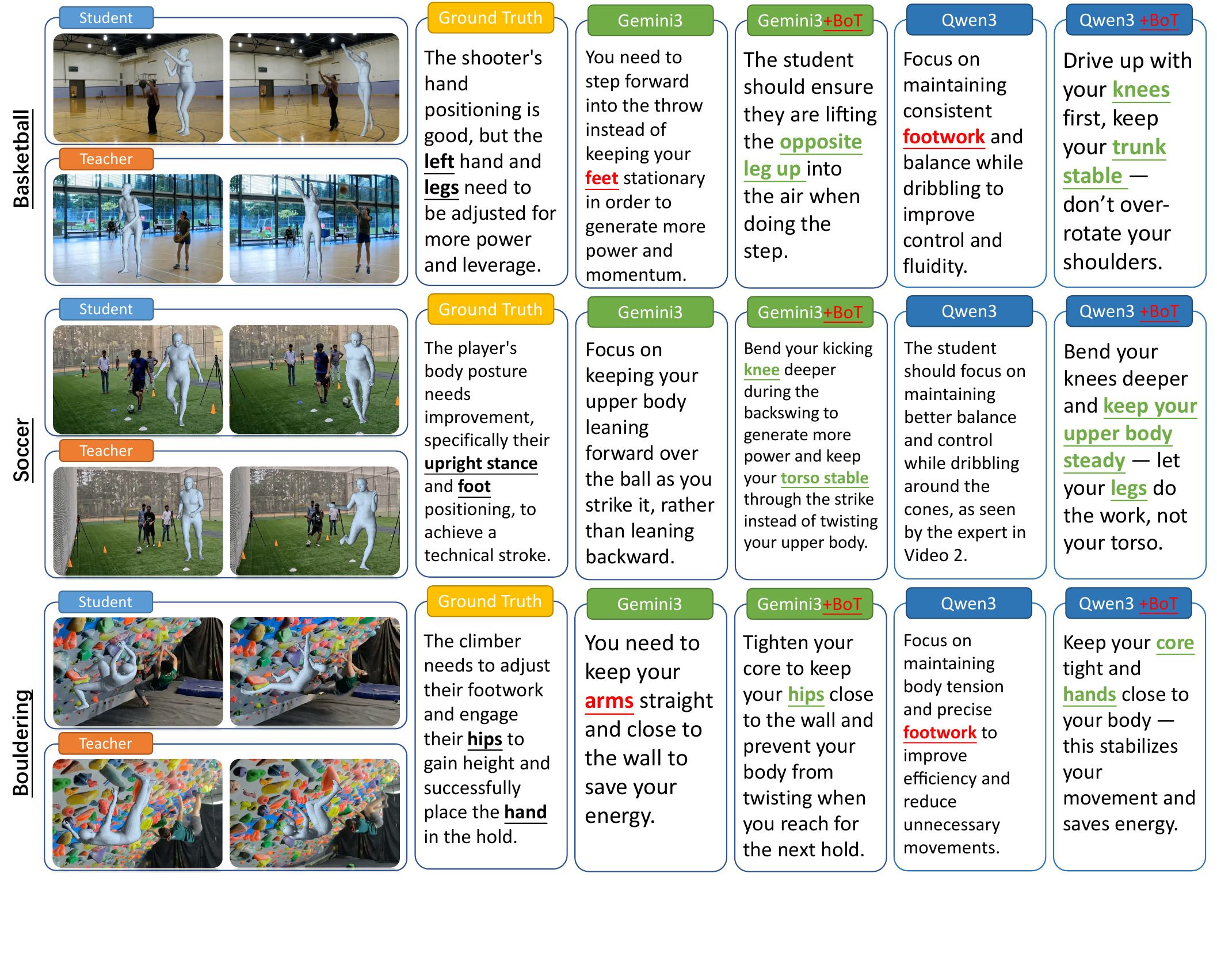}    
    \label{fig:visquality}
    \vspace{-20pt}
\end{figure}

\begin{table}[htbp]
\centering
\caption{Robustness Analysis of BoT-Feedback across Varying ($\sigma \in \{0, 0.05, 0.1\}$) Skeleton Noise Levels, Biomechanics of Thought Step 1 Random Action, and Multi-view Skeleton Extraction.}
\label{tab:noise_robustness_reorder}
\resizebox{\columnwidth}{!}{%
\begin{tabular}{@{}l c c c c c@{}}
\toprule
\multirow{2}{*}{\textbf{Action}} & \multirow{2}{*}{\textbf{Baseline}} & \multirow{2}{*}{\makecell{\textbf{Skeleton Noise} \\ \textbf{0.05 / 0.1}}} & \multirow{2}{*}{\makecell{\textbf{Random} \\ \textbf{Action}}} & \multirow{2}{*}{\textbf{Multiview}} & \multirow{2}{*}{\textbf{Ours}} \\
 & & & & & \\ \midrule
SP  & 1.61 & 2.71 / 2.71 & 2.66 & 2.93 & 3.56 \\
SD  & 2.38 & 2.89 / 2.95 & 2.77 & 3.02 & 2.99 \\
BRL & 1.77 & 2.75 / 2.72 & 2.80 & 3.02 & 3.15 \\
BM  & 1.86 & 3.10 / 2.99 & 2.89 & 3.06 & 2.99 \\
BJS & 1.98 & 2.98 / 3.05 & 2.81 & 3.00 & 3.22 \\
BO  & 2.07 & 2.36 / 2.20 & 2.42 & 2.84 & 2.50 \\ \midrule
\textbf{Average} & \textbf{1.95} & \textbf{2.80 / 2.77} & \textbf{2.73} & \textbf{2.98} & \textbf{3.07} \\ \bottomrule
\end{tabular}%
}
\vspace{-15pt}
\end{table}

\subsubsection{Robustness Analysis under Skeleton and Action Identification Noise}
To evaluate the robustness of BoT-Feedback against pose bias or identification failure, we analysed its performance under Gaussian noise injected into the 3D skeleton and random action assignments during Biomechanics of Thought. As shown in \Cref{tab:noise_robustness_reorder}, our framework consistently outperforms the baseline (w/o BoT-Feedback). Even under high-intensity skeleton noise ($\sigma=0.1$), the average score (2.77) surpasses the noise-free baseline (1.95) by 42\%, showing remarkable stability against joint-level jitter during pose estimation. Additionally, to simulate inaccurate action identification, we replaced the first-stage results of Biomechanics of Thought with random actions. The final score (2.73) still leads the baseline by 40\%. This confirms that our dual-video contrastive mechanism and multi-step reasoning effectively rectify low-quality inputs and semantic deviations, ensuring robustness. To further demonstrate robustness to skeleton pose noise, we extend our evaluation to multi-view skeleton extraction, where 3D skeletons are estimated from three exocentric camera views and aggregated via cross-view aggregation and temporal smoothing. Despite the significantly higher acquisition cost of multi-view capture, the resulting performance remains comparable to the single-view setting (-0.09). Overall, the limited degradation under severe pose noise or improved pose indicates that BoT relies on consistent biomechanical trends rather than precise joint localisation, suggesting that the reasoning framework remains effective even when upstream pose estimation is imperfect.

\subsubsection{Qualitative Results}
To qualitatively demonstrate the effectiveness of our framework, \Cref{fig:visquality} presents the teacher-student video pairs alongside their corresponding 3D mesh and generated feedback. The results reveal that MLLMs frequently produce generic or incorrect output (e.g., vague suggestions such as "maintain consistent footwork and balance"). In contrast, by integrating our BoT-Feedback, the models identify actions and critical body segments from underlying skeleton data. By analysing these biomechanical features, the framework bridges the gap from low-level 2D pixels to high-level semantic physical mechanics, mitigating common issues such as physical hallucinations and generic outputs while providing precise, explainable professional guidance (e.g., specific joint extensions and limb positioning). These outputs align closely with the ground truth expert feedback. The evidence consistently indicates that the performance gains arise from changing the reasoning space—from latent semantic reasoning to explicit biomechanical reasoning—rather than from simply providing additional information or prompt engineering.

\section{Conclusion and Future Work}
We introduce BoT-Feedback, which grounds MLLM reasoning in structured biomechanical evidence through Biomechanics of Thought. Across twelve MLLMs, the framework consistently improves feedback quality, interpretability and robustness while reducing biomechanical hallucinations. Future work will investigate tighter integration of biomechanical representations into multimodal reasoning, and extend BoT-Feedback to broader action categories and real-world coaching applications.


\bibliography{aaai2027}

@article{barrios2010gait,
  title={Gait retraining to reduce the knee adduction moment through real-time visual feedback of dynamic knee alignment},
  author={Barrios, Joaquin A and Crossley, Kay M and Davis, Irene S},
  journal={Journal of Biomechanics},
  volume={43},
  number={11},
  pages={2208--2212},
  year={2010},
  publisher={Elsevier}
}

@article{hewett2005biomechanical,
  title={Biomechanical measures of neuromuscular control and valgus loading of the knee predict anterior cruciate ligament injury risk in female athletes: a prospective study},
  author={Hewett, Timothy E and Myer et al.},
  journal={The American Journal of Sports Medicine},
  pages={492--501},
  year={2005},
  publisher={Sage Publications Sage CA: Thousand Oaks, CA}
}

@InProceedings{grauman2024egoexo4dunderstandingskilledhuman,
      title={Ego-Exo4D: Understanding Skilled Human Activity from First- and Third-Person Perspectives},
  author={Ego-Exo4D Team},
  booktitle={Proceedings of the IEEE/CVF Conference on Computer Vision and Pattern Recognition (CVPR)},
  year={2024}
}

@inproceedings{ashutosh2025expertafexpertactionablefeedback,
      title={ExpertAF: Expert Actionable Feedback from Video}, 
      author={Kumar Ashutosh and Tushar Nagarajan and Georgios Pavlakos and Kris Kitani and Kristen Grauman},
      year={2025},
    booktitle={Proceedings of the IEEE/CVF Conference on Computer Vision and Pattern Recognition (CVPR)},

}

@inproceedings{li2023blip2bootstrappinglanguageimagepretraining,
  title={BLIP-2: Bootstrapping Language-Image Pre-training with Frozen Image Encoders and Large Language Models},
  author={Li, Junnan and Li, Dongxu and Savarese, Silvio and Hoi, Steven},
  booktitle={Proceedings of the 40th International Conference on Machine Learning (ICML)},
  year={2023}
}

@inproceedings{liu2023visualinstructiontuning,
  title={Visual Instruction Tuning},
  author={Liu, Haotian and Li, Chunyuan and Wu, Qingyang and Lee, Yong Jae},
  booktitle={Proceedings of the 37th International Conference on Neural Information Processing Systems (NeurIPS)},
  year={2023}
}

@inproceedings{zhu2023minigpt4enhancingvisionlanguageunderstanding,
title={MiniGPT-4: Enhancing Vision-Language Understanding with Advanced Large Language Models},
  author={Zhu, Deyao and Chen, Jun and Shen, Xiaoqian and Li, Xiang and Elhoseiny, Mohamed},
  booktitle={Proceedings of the 12th International Conference on Learning Representations (ICLR)},
  year={2024}
}

@inproceedings{chen2024internvlscalingvisionfoundation,
  title={InternVL: Scaling up Vision Foundation Models and Aligning for Generic Visual-Linguistic Tasks},
  author={InternVL Team},
  booktitle={Proceedings of the IEEE/CVF Conference on Computer Vision and Pattern Recognition (CVPR)},
  year={2024}
}

@inproceedings{li2024llavanextinterleavetacklingmultiimagevideo,
  title={LLaVA-NeXT-Interleave: Tackling Multi-image, Video, and 3D in Large Multimodal Models},
  author={Li, Feng and Zhang, Renrui and Zhang, Hao and Zhang, Yuanhan and Li, Bo and Li, Wei ox and Ma, Zejun and Li, Chunyuan},
  booktitle={arXiv preprint arXiv:2407.07895},
  year={2024},
  note   = {Last Accessed: 2026-07-24}
}

@inproceedings{pirsiavash2014assessing,
  title={Assessing the quality of actions},
  author={Pirsiavash, Hamed and Vondrick, Carl and Torralba, Antonio},
  booktitle={Proceedings of the European Conference on Computer Vision (ECCV)},
  pages={556--571},
  year={2014}
  
}

@inproceedings{parmar2019action,
  title={Action quality assessment across multiple actions},
  author={Parmar, Paritosh and Morris, Brendan},
  booktitle={Proceedings of the IEEE winter conference on applications of computer vision (WACV)},
  pages={1468--1476},
  year={2019},
}

@inproceedings {parmar2017learning,
author = {P. Parmar and B. Morris},
booktitle = {IEEE Conference on Computer Vision and Pattern Recognition Workshops (CVPRW)},
title = {Learning to Score Olympic Events},
year = {2017},
pages = {76-84}
}

@article{xu2019learning,
  title={Learning to score figure skating sport videos},
  author={Xu, Chengming and Fu, Yanwei and Zhang, Bing and Chen, Zitian and Jiang, Yu-Gang and Xue, Xiangyang},
  journal={IEEE transactions on circuits and systems for video technology (TCSVT)},
  volume={30},
  number={12},
  pages={4578--4590},
  year={2019},
  publisher={IEEE}
}

@inproceedings{xiang2018s3d,
  title={S3d: Stacking segmental p3d for action quality assessment},
  author={Xiang, Xiang and Tian, Ye and Reiter, Austin and Hager, Gregory D and Tran, Trac D},
  booktitle={25th IEEE International Conference on Image Processing (ICIP)},
  pages={928--932},
  year={2018},
  organization={IEEE}
}

@inproceedings{xu2022likert,
  title={Likert scoring with grade decoupling for long-term action assessment},
  author={Xu, Angchi and Zeng, Ling-An and Zheng, Wei-Shi},
  booktitle={Proceedings of the IEEE/CVF Conference on Computer Vision and Pattern Recognition (CVPR)},
  pages={3232--3241},
  year={2022}
}

@InProceedings{Doughty_2019_CVPR,
author = {Doughty, Hazel and Mayol-Cuevas, Walterio and Damen, Dima},
title = {The Pros and Cons: Rank-Aware Temporal Attention for Skill Determination in Long Videos},
booktitle = {Proceedings of the IEEE/CVF Conference on Computer Vision and Pattern Recognition (CVPR)},
year = {2019}
}

@inproceedings{xu2024fineparser,
  title={FineParser: A Fine-grained Spatio-temporal Action Parser for Human-centric Action Quality Assessment},
  author={Xu, Jinglin and Yin, Sibo and Zhao, Guohao and Wang, Zishuo and Peng, Yuxin},
  booktitle={Proceedings of the IEEE/CVF Conference on Computer Vision and Pattern Recognition (CVPR)},
  pages={14628--14637},
  year={2024}
}

@inproceedings{xu2025languageguided,
  title={Language-Guided Audio-Visual Learning for Long-Term Sports Assessment},
  author={Xu, Huangbiao and Ke, Xiao and Wu, Huanqi and Xu, Rui and Li, Yuezhou and Guo, Wenzhong},
  year={2025},
  booktitle={In Proceedings of the IEEE/CVF Conference on Computer Vision and Pattern Recognition (CVPR)},
  pages={23967--23977}
}

@inproceedings{Zhou_2024,
   title={CoFInAl: Enhancing Action Quality Assessment with Coarse-to-Fine Instruction Alignment},
   booktitle={Proceedings of the Thirty-Third International Joint Conference on Artificial Intelligence (IJCAI)},
   author={Zhou, Kanglei and Li, Junlin and Cai, Ruizhi and Wang, Liyuan and Zhang, Xingxing and Liang, Xiaohui},
   year={2024},
    pages={1771–1779},
    }

@article{dong2026uila,
  author       = {Dong, Xu and Liu, Xinran and Li, Wanqing and Adeyemi-Ejeye, Anthony and Gilbert, Andrew},
  title        = {UIL-AQA: Uncertainty-Aware Clip-Level Interpretable Action Quality Assessment},
  journal      = {International Journal of Computer Vision (IJCV)},
  year         = {2026},
  volume       = {134},
  number       = {24}}

@article{hou2025interpretable,
  author       = {Hou, Shouming and Xia, Aoyu and Lu, Zixuan and Yang, Weibo},
  title        = {Interpretable Two-Stage Action Quality Assessment Via 3D Human Pose Estimation and Dynamic Feature Alignment},
  journal      = {The Visual Computer},
  year         = {2025},
  pages        = {11463--11478},
}

@inproceedings{han2025finecausal,
  author       = {Han, Ruisheng and Zhou, Kanglei and Atapour-Abarghouei, Amir and Liang, Xiaohui and Shum, Hubert P. H.},
  title        = {Finecausal: A Causal-Based Framework for Interpretable Fine-Grained Action Quality Assessment},
  booktitle    = {IEEE/CVF Conference on Computer Vision and Pattern Recognition Workshops  (CVPRW)},
  year         = {2025},
  pages        = {6008--6017},
}

@article{qi2025explainable,
  author       = {Qi, Mengshi and Wu, Yeteng and Zhang, Xianlin and Ma, Huadong},
  title        = {Explainable Action Form Assessment by Exploiting Multimodal Chain-of-Thoughts Reasoning},
  journal      = {arXiv preprint arXiv:2512.15153},
  year         = {2025},
  note   = {Last Accessed: 2026-07-24}
}

@article{xu2025qualityguided,
  author       = {Xu, Huangbiao and Wu, Huanqi and Ke, Xiao and Li, Yuezhou and Xu, Rui and Guo, Wenzhong},
  title        = {Quality-Guided Vision-Language Learning for Long-Term Action Quality Assessment},
  journal      = {IEEE Transactions on Multimedia (TMM)},
  year         = {2025},
  volume       = {27},
  pages        = {7326--7339},
}

@inproceedings{wei2023chainofthoughtpromptingelicitsreasoning,
  title={Chain-of-Thought Prompting Elicits Reasoning in Large Language Models},
  author={Wei, Jason and Wang, Xuezhi and Schuurmans et al.},
  booktitle={Proceedings of the 36th International Conference on Neural Information Processing Systems (NeurIPS)},
  year={2022}
}

@inproceedings{treeofthought,
  title={Tree of Thoughts: Deliberate Problem Solving with Large Language Models},
  author={Shunyu Yao and Dian Yu and Jeffrey Zhao and Izhak Shafran and Thomas L. Griffiths and Yuan Cao and Karthik Narasimhan},
  booktitle={Proceedings of the 37th International Conference on Neural Information Processing Systems (NeurIPS)},
  year={2023}
}

@inproceedings{Besta_2024,
  title={Graph of Thoughts: Solving Elaborate Problems with Large Language Models},
  author={Besta, Maciej and Blach, Nils and Kubicek, Ales and Gerstenberger et al.},
  booktitle={Proceedings of the 38th AAAI Conference on Artificial Intelligence (AAAI)},
  year={2024}
}

@inproceedings{chen2023programthoughtspromptingdisentangling,
  title={Program of Thoughts Prompting: Disentangling Computation from Reasoning for Numerical Reasoning Tasks},
  author={Chen, Wenhu and Ma, Xueguang and Wang, Xinyi and Cohen, William W.},
  booktitle={Transactions on Machine Learning Research (TMLR)},
  year={2023}
}

@inproceedings{fei2024videoofthoughtstepbystepvideoreasoning,
  title={Video-of-Thought: Step-by-Step Video Reasoning from Perception to Cognition},
  author={Fei, Hao and Wu, Shengqiong and Ji, Wei and Zhang, Hanwang and Zhang, Meishan and Lee, Mong-Li and Hsu, Wynne},
  booktitle={Proceedings of the 41st International Conference on Machine Learning (ICML)},
  year={2024}
}

@inproceedings{shimada2020physcapphysicallyplausiblemonocular,
  title={PhysCap: Physically Plausible Monocular 3D Motion Capture in Real Time},
  author={Shimada, Soshi and Golyanik, Vladislav and Xu, Weipeng and Theobalt, Christian},
  booktitle={Proceedings of the ACM SIGGRAPH Asia Conference (SIGGRAPH Asia)},
  year={2020}
}

@inproceedings{vonmarcard2017sparseinertialposerautomatic,
  title={Sparse Inertial Poser: Automatic 3D Human Pose Estimation from Sparse IMUs},
  author={von Marcard, Timo and Rosenhahn, Bodo and Black, Michael J. and Pons-Moll, Gerard},
  booktitle={Proceedings of the Eurographics (EG)},
  year={2017}
}

@InProceedings{Huang_2024_CVPR,
    author    = {Huang, Buzhen and Li, Chen and Xu, Chongyang and Pan, Liang and Wang, Yangang and Lee, Gim Hee},
    title     = {Closely Interactive Human Reconstruction with Proxemics and Physics-Guided Adaption},
    booktitle = {Proceedings of the IEEE/CVF Conference on Computer Vision and Pattern Recognition (CVPR)},
    month     = {June},
    year      = {2024},
    pages     = {1011-1021}
}

@misc{sam3dteam2025sam3d3dfyimages,
  title={SAM 3D: 3Dfy Anything in Images},
  author={SAM 3D Team},
  booktitle={arXiv preprint arXiv:2511.16624},
  year={2025},
  note   = {Last Accessed: 2026-07-24}
}

@inproceedings{1964AnaCh,
  title={Smoothing and differentiation of data by simplified least squares procedures},
  author={Savitzky, Abraham and Golay, Marcel J. E.},
  booktitle={Analytical Chemistry},
  year = 1964,
   volume = 36,
}

@inproceedings{lin2023video,
  title={Video-LLaVA: Learning United Visual Representation by Alignment Before Projection},
  author={Video-LLaVA team},
  booktitle={Proceedings of the 2024 Conference on Empirical Methods in Natural Language Processing (EMNLP)},
  year={2024}
}

@inproceedings{maaz2023video,
  title={Video-ChatGPT: Towards Detailed Video Understanding via Large Vision and Language Models},
  author={Maaz, Muhammad and Rasheed, Hanoona and Khan, Salman and Khan, Fahad},
  booktitle={Proceedings of the 62nd Annual Meeting of the Association for Computational Linguistics (ACL)},
  year={2024},
  pages={12585--12602}
}

@inproceedings{chen2024internvl2,
  title={InternVL2.5: Scaling Progressively for Better Multimodal Understanding},
  author={Chen, Zhe and Wang, Weiyuan and Cao et al.},
  booktitle={arXiv preprint arXiv:2412.05271},
  year={2024},
note   = {Last Accessed: 2026-07-24}
}

@inproceedings{lu2024deepseek,
  title={DeepSeek-VL2: Mixture-of-Experts Vision-Language Models for Advanced Understanding},
  author={Lu, Haoyu and Yang, Wenjun and Zhang, Zhenda and others},
  booktitle={arXiv preprint arXiv:2412.13602},
  year={2024},
  note   = {Last Accessed: 2026-07-24}
}

@inproceedings{damon2024videollama3,
  title={Video-LLaMA 3: Frontier Language Models for Video Understanding},
  author={Damon, Fan and Zhang, Yuqi and others},
  booktitle={arXiv preprint arXiv:2412.21057},
  year={2024},
  note   = {Last Accessed: 2026-07-24}
}

@inproceedings{li2025videochatr1,
  title={VideoChat-R1: Towards Real-Time and Robust Video Understanding},
  author={Li, KunChang and Wang, Jiashuo and others},
  booktitle={arXiv preprint arXiv:2502.13110},
  year={2025},
  note   = {Last Accessed: 2026-07-24}
}

@inproceedings{qwen3vl2025,
  title={Qwen3-VL Technical Report},
  author={Qwen Team},
  booktitle={arXiv preprint arXiv:2511.21631},
  year={2025},
  note   = {Last Accessed: 2026-07-24}
}

@inproceedings{wang2025internvideo25empoweringvideomllms,
  title={InternVideo2.5: Empowering Video MLLMs with Long and Rich Context Modeling},
  author={Wang, Yi and Li, Xinhao and Yan, Ziang and He et al.},
  booktitle={arXiv preprint arXiv:2501.12386},
  year={2025},
  note   = {Last Accessed: 2026-07-24}
}

@inproceedings{winter2009biomechanics,
  title={Biomechanics and Motor Control of Human Movement},
  author={Winter, David A.},
  booktitle={Biomechanics and Motor Control of Human Movement (4th Edition)},
  year={2009}
}

@inproceedings{mcpherson1989relation,
  title={Relation of knowledge and strategies to creative and typical performance in tennis},
  author={McPherson, Sue L. and Thomas, Jerry R.},
  booktitle={Journal of Teaching in Physical Education (JTPE)},
  year={1989},
  volume={8},
  pages={139--149}
}

@inproceedings{Sakoe1978DynamicPA,
  title={Dynamic programming algorithm optimization for spoken word recognition},
  author={Sakoe, Hiroaki},
  booktitle={IEEE Transactions on Acoustics, Speech, and Signal Processing (TASSP)},
  year={1978},
  volume={26},
  pages={159--165}
}

@inproceedings{comanici2025gemini25pushingfrontier,
  title={Gemini 2.5: Pushing the Frontier with Advanced Reasoning, Multimodality, Long Context, and Next Generation Agentic Capabilities},
  author={Gemini 2.5 Team},
  booktitle={arXiv preprint arXiv:2507.06261},
  year={2025},
  note   = {Last Accessed: 2026-07-24}
}

@inproceedings{cuturi2018softdtwdifferentiablelossfunction,
  title={Soft-DTW: a Differentiable Loss Function for Time-Series},
  author={Cuturi, Marco and Blondel, Mathieu},
  booktitle={Proceedings of the 34th International Conference on Machine Learning (ICML)},
  year={2017}
}

@inproceedings{yang2026sam3dbodyrobust,
  title={SAM 3D Body: Robust Full-Body Human Mesh Recovery},
  author={SAM 3D Body Team},
  booktitle={arXiv preprint arXiv:2602.15989},
  year={2026},
  note   = {Last Accessed: 2026-07-24}
}

@inproceedings{team2026gemini3,
  title={Gemini 3: Next-Generation Multimodal Models with Enhanced Reasoning and Native Audio-Visual Understanding},
  author={Gemini Team, Google},
  booktitle={arXiv preprint arXiv:2602.12844},
  year={2026},
  note   = {Last Accessed: 2026-07-24}
}

@misc{openai2024gpt4o,
      title={GPT-4 Technical Report}, 
      author={GPT-4 Team},
      year={2024},
      booktitle={arXiv preprint arXiv:2303.08774},
      note   = {Last Accessed: 2026-07-24}
}

@inproceedings{liu2023gevalnlgevaluationusing,
  title={G-Eval: NLG Evaluation using GPT-4 with Better Human Alignment},
  author={Liu, Yang and Iter, Dan and Xu, Yichong and Wang, Shuohang and Xu, Ruochen and Zhu, Chenguang},
  booktitle={Proceedings of the 2023 Conference on Empirical Methods in Natural Language Processing (EMNLP)},
  year={2023},
  pages={2511--2522}
}

@inproceedings{zhou2025physvlmenablingvisuallanguage,
  title={PhysVLM: Enabling Visual Language Models to Understand Robotic Physical Reachability},
  author={Zhou, Weijie and Tao, Manli and Zhao et al.},
  booktitle={arXiv preprint arXiv:2503.08481},
  year={2025},
  note   = {Last Accessed: 2026-07-24}
}

@inproceedings{keogh2001derivative,
  title={Derivative dynamic time warping},
  author={Keogh, Eamonn J. and Pazzani, Michael J.},
  booktitle={Proceedings of the First SIAM International Conference on Data Mining (SDM)},
  year={2001},
  pages={1--11}
}

@inproceedings{stechly2025chainthoughtlessnessanalysiscot,
  title={Chain of Thoughtlessness? An Analysis of CoT in Planning},
  author={Stechly, Kaya and Valmeekam, Karthik and Kambhampati, Subbarao},
  booktitle={Advances in Neural Information Processing Systems (NeurIPS)},
  year={2024},
  pages={29841--29853}
}

@inproceedings{barez2025chain,
  title={Chain-of-thought is not explainability},
  author={Barez, Fazl and Wu, Tung-Yu and Arcuschin et al.},
  booktitle={arXiv preprint arXiv:2501.04364},
  year={2025},
  note   = {Last Accessed: 2026-07-24}
}

@article{lees2002technique,
  title={Technique analysis in sports: a critical review},
  author={Lees, Adrian},
  journal={Journal of sports sciences},
  pages={813--828},
  year={2002},
}

@book{simon1978information,
  title={Information-processing theory of human problem solving},
  author={Simon, Herbert A},
  year={1978},
  publisher={Erlbaum Hillsdale, NJ}
}


\end{document}